\documentclass[conference]{IEEEtran}
\IEEEoverridecommandlockouts
\usepackage{cite}
\usepackage{amsmath,amssymb,amsfonts}
\usepackage{algorithmic}
\usepackage{graphicx}
\usepackage{textcomp}
\usepackage{xcolor}
\usepackage{bm}
\usepackage{fancyhdr}
\fancypagestyle{plain}{\pagestyle{fancy}}

\def\BibTeX{{\rm B\kern-.05em{\sc i\kern-.025em b}\kern-.08em
    T\kern-.1667em\lower.7ex\hbox{E}\kern-.125emX}}
\begin{document}

\fancyhead[CO,CE]{IEEE Copyright Notice Copyright (c) 2023 IEEE Personal
use of this material is permitted. 
Accepted to be published in 2022 IEEE International Conference on Big Data (Big Data).
DOI: 10.1109/BigData55660.2022.10020236}

\title{QUBO-inspired Molecular Fingerprint for Chemical Property Prediction}

\makeatletter
\newcommand{\linebreakand}{%
  \end{@IEEEauthorhalign}
  \hfill\mbox{}\par
  \mbox{}\hfill\begin{@IEEEauthorhalign}
}
\makeatother

\ifthenelse{1>0}{
\author{
\IEEEauthorblockN{Koichiro Yawata}
\IEEEauthorblockA{\textit{Research and Development Group} 
\textit{Hitachi Ltd.}\\
Kokubunji-shi, Japan \\
koichiro.yawata.rt@hitachi.com}
\and
\IEEEauthorblockN{Yoshihiro Osakabe}
\IEEEauthorblockA{\textit{Research and Development Group} 
\textit{Hitachi Ltd.}\\
Kokubunji-shi, Japan \\
yoshihiro.osakabe.fj@hitachi.com}

\linebreakand 
\IEEEauthorblockN{Takuya Okuyama}
\IEEEauthorblockA{\textit{Research and Development Group} 
\textit{Hitachi Ltd.}\\
Kokubunji-shi, Japan \\
takuya.okuyama.mn@hitachi.com}
\and
\IEEEauthorblockN{Akinori Asahara}
\IEEEauthorblockA{\textit{Research and Development Group} 
\textit{Hitachi Ltd.}\\
Kokubunji-shi, Japan \\
akinori.asahara.bq@hitachi.com}
}
}
{
  \author{
    \IEEEauthorblockN{1\textsuperscript{st} Anonymous}
    \IEEEauthorblockA{\textit{Anonymous: dept. name of organization} 
    \textit{.}\\
    City, Country \\
    email address or ORCID}
    \and
    \IEEEauthorblockN{2\textsuperscript{nd} Anonymous}
    \IEEEauthorblockA{\textit{Anonymous: dept. name of organization} 
    \textit{.}\\
    City, Country \\
    email address or ORCID}
    \linebreakand 
    \IEEEauthorblockN{3\textsuperscript{rd} Anonymous}
    \IEEEauthorblockA{\textit{Anonymous: dept. name of organization} 
    \textit{.}\\
    City, Country \\
    email address or ORCID}
    \and
    \IEEEauthorblockN{4\textsuperscript{th} Anonymous}
    \IEEEauthorblockA{\textit{Anonymous: dept. name of organization} 
    \textit{.}\\
    City, Country \\
    email address or ORCID}
  }
}

\maketitle

\begin{abstract}
Molecular fingerprints are widely used for predicting chemical properties,
and selecting appropriate fingerprints is important.
We generate new fingerprints
based on the assumption that a performance of prediction
using a more effective fingerprint is better.
We generate effective interaction fingerprints
that are the product of multiple base fingerprints.
It is difficult to evaluate all combinations of interaction fingerprints
because of computational limitations.
Against this problem,
we transform a problem of searching more effective interaction fingerprints
into a quadratic unconstrained binary optimization problem.
In this study, we found effective interaction fingerprints using QM9 dataset.
\end{abstract}


\section{Introduction}
Predicting chemical properties using machine learning is
one of the most popular topics in the material informatics.
Molecular fingerprints are widely used
because they can describe molecular structure as tabular data \cite{mauri2020}.
Many fingerprints have been proposed,
and selecting appropriate fingerprints is important from them \cite{durant2002}.
However, the factor of determining the chemical properties is complicated.
Even if crucial factor is described by multiple fingerprints,
treating them independently is not effective in terms of machine learning.
In fact, interaction features that are the product of multiple features,
are used as a feature engineering technique \cite{zheng2018}.

We generate new fingerprints on the assumption that
a performance of prediction based on a more effective fingerprint is better.
The performance of each fingerprint is evaluated by
the prediction error of chemical properties based on splitting training samples using each fingerprint
like decision trees with depth 1.
Then we define interaction fingerprints.
They are fingerprints that are described by the product of multiple base fingerprints,
and we try to search effective fingerprints from them.
We use existing fingerprints as base fingerprints.
Examples of interaction fingerprints
and molecules satisfying the interaction fingerprint are shown in Fig. \ref{fig:Bigdata4}.
In this study, MACCS keys fingerprint \cite{durant2002} is used as base fingerprints,
and the notation of each base fingerprint is based on RDKit \cite{rdkit}.

However, it is difficult to evaluate all combinations of interaction fingerprints
because of computational limitations.
Against this problem,
we transform a problem of searching more effective interaction fingerprints
into a quadratic unconstrained binary optimization (QUBO) problem,
which is called QUBO decision trees \cite{yawata2022}.
QUBO problem can be solve approximately faster by an annealing machine \cite{kadowaki1998quantum}.
As a result, by using QUBO decision trees,
it is expected to obtain effective interaction fingerprints.

\begin{figure}
  \centering
  \includegraphics[width=0.9\linewidth,page=3]{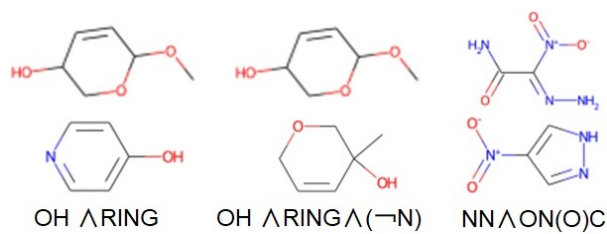}
  \caption{Examples of interaction fingerprints using multiple MACCS keys fingerprint.
OH$\land$RING$\land$($\lnot$N) indicates interaction fingerprints means that
molecules are with hydroxy group and ring structure and without nitrogen atom.
  }\label{fig:Bigdata4}
\end{figure}

\section{Related Work}
\textbf{Annealing machines}.
There has been a lot of research on annealing machines, both in hardware and software.
They have strengths in solving combinatorial optimization problems
and have been used in the real world \cite{stollenwerk2019quantum}.
Quantum annealing \cite{kadowaki1998quantum} and the adiabatic quantum evolution 
algorithm \cite{farhi2001quantum} drove the development of annealing machines.
D-Wave provides a quantum annealing machine in the cloud \cite{shin2014quantum}.
Annealing machines that use GPUs with a technique called momentum annealing
have also been introduced \cite{okuyama2019binary, tao2020}.

\textbf{Machine learning optimized as QUBO problem}.
There have been many attempts to solve problems with annealing technology
by formulating it as a QUBO problem for machine learning algorithms.
In \cite{kurihara2009quantum}, annealing machines have been used for clustering.
In \cite{prasanna2021qubo}, three methods, linear regression, a support vector machine (SVM), and
balanced k-means clustering were transformed into QUBO problems.
Annealing machines have also been used in deep learning
\cite{sasdelli2021quantum, rere2015simulated, sato2019},
image processing \cite{birdal2021quantum,li2020quantum,cruz2018qubo}
and Bayesian network \cite{ogorman2015, shikuri2020}.

\section{Problem Setup}
\label{problemsetup}
Let $\bm{X}_{i,j}$ be the value of base fingerprints $j$,
$t^{(i)}$ be the target value for sample $i$,
$y^{(i)}$ be the predicted value for sample $i$,
and $g^{(i)}$ be the value of generated fingerprint for sample $i$.
$g^{(i)}$ can be written as follow:
\begin{equation}
  g^{(i)} = \prod_{f_j=1, j \in \mathbb{F}} \bm{X}_{i,j},
\end{equation} 
where $\mathbb{F}$ is the set of base fingerprints
and $f$ is the fingerprint vector.
$f_j=1$ means that generated interaction fingerprint uses fingerprint $j$,
and such fingerprints are called producted fingerprints in this study.
In the first column of Table \ref{tab:tab5},
The generated interaction fingerprint is OH$\land$RING,
the base fingerprints are OH, $\lnot$OH, RING, $\lnot$RING, N and $\lnot$N,
and the producted fingerprints are OH and RING.
Let $N_\mathbb{F}$ be the number of prepared fingerprints.
Then the number of fingerprints used for interaction fingerprints $U$
can be written as follow:
\begin{equation*}
  U = \sum_{j \in \mathbb{F}} f_j.
\end{equation*}
Because there are $2^{N_\mathbb{F}}$ different combinations of fingerprints,
it is difficult to search for the optimal solution in general.
Therefore, in this study we use annealing machines,
which have strengths in solving combinatorial optimization problems.
Furthermore, let $\mathbb{S}$ be the set of all samples,
$\mathbb{S}_1$ be the set of samples that satisfied $g^{(i)}=1$,
and $\mathbb{S}_0$ be the set of samples that did not.
The number of samples in $\mathbb{S}$ is denoted by $N_\mathbb{S}$.

\begin{table}
  \centering
  \caption{Interaction Fingerprint and Fingerprint Vector.}\label{tab:tab5}
  \begin{tabular}{|c|cccccc|c|}
    \hline
\begin{tabular}{c}
  Interaction\\
  Fingerprint
  \end{tabular}&OH&$\lnot$OH&RING&$\lnot$RING&N&$\lnot$N&U \\
     \hline
    OH$\land$RING&1&0&1&0&0&0&2\\
    OH$\land$RING$\land(\lnot$N)&1&0&1&0&0&1&3\\
    \hline
  \end{tabular}
\end{table}

\textbf{QUBO problem}.
In an annealing machine, the parameters to be optimized are represented
by QUBO variables with binary values $\theta_l\in\{0,1\}$,
and the objective variable is represented by the Hamiltonian of the QUBO form:
\begin{equation}
    H=-\sum_{l<m}Q_{lm}\theta_l\theta_m-\sum_lb_l\theta_l, \label{ising}
\end{equation} 
where the $Q_{lm}$ and $b_l$ are coefficients that characterize the QUBO problem.

The QUBO problem is formulated by expressing the Hamiltonian
as the sum of a loss function $L$ and a constraint function $C$
via the QUBO format, as follow:
\begin{equation*}
  H=L\left(\boldsymbol{X},\boldsymbol{t}|\boldsymbol{\theta}\right)+C(\boldsymbol{\theta})
\end{equation*}
Here, $\boldsymbol{\theta}$ is a binary vector representing the model parameters.
$\boldsymbol{X}$ is the explanatory variable for sample $i$,
and $\boldsymbol{t}$ is a vector representing the target variable.
In addition, $L$ is the optimization target, and $C(\boldsymbol{\theta})$ is a set to obtain a valid solution,
which is generally required to be $C(\boldsymbol{\theta})=0$.
The annealing machine is used to find $\boldsymbol{\theta}$ such that the Hamiltonian is minimized.

\section{QUBO Decision Tree}
The mean squared error (MSE) is often used to learn decision trees.
The MSE in a decision tree for regression can be written as follows:
\begin{equation*}
  \text{MSE}=\frac{1}{N_\mathbb{S}}(\sum_{{i}\in S_1}\left({\text{pred}}_1-t^{(i)}\right)^2+
  \sum_{{i}\in \mathbb{S}_0}\left({\text{pred}}_0-t^{(i)}\right)^2),
\end{equation*}
where $\text{pred}_1$ is the estimated value when the interaction fingerprints is satisfied,
and $\text{pred}_0$ is the estimated value when it is not satisfied.
Learning in decision trees for regression involves finding the interaction fingerprints for splitting
and the $\text{pred}_1,\text{pred}_0$ to reduce the MSE.
It is obvious that $\text{pred}_1$ and $\text{pred}_0$ are the means of the samples divided by the interaction fingerprints.
Therefore, the MSE is equal to the sum of the variance of the split sample groups
weighted by the proportions of the sample groups.
Accordingly, the MSE can be written as the following equation:
\begin{align*}
  \text{MSE}=&\sum_{b=0,1}{\text{Var}\left(\left\{t^{(i)}\middle| i\in \mathbb{S}_b\right\}\right)\frac{N_{\mathbb{S}_b}}{N_\mathbb{S}}}\\
  =&\sum_{b=0,1}{(\frac{1}{N_\mathbb{S}}\sum_{i\in \mathbb{S}_b} t^{(i)2}-\frac{1}{N_\mathbb{S}N_{\mathbb{S}_b}}(\sum_{i\in \mathbb{S}_b} t^{(i)})^2)}.
\end{align*}
However, the MSE is difficult to formulate in a QUBO problem because
it involves division using the variable $N_{\mathbb{S}_b}$ that is the sum of QUBO variables.
Here, because $N_\mathbb{S}$ is the total number of samples and thus a constant,
this is not a problem.
Hence, we propose the square weighted MSE (SWMSE).
In the SWMSE, instead of using the proportion of the sample group as the weight in calculating the sum of the variances,
the square of that value is used:
\begin{align}
  \text{SWMSE}=&\sum_{b=0,1}{\text{Var}\left(\left\{t^{(i)}\middle| i\in \mathbb{S}_b\right\}\right)\left(\frac{N_{\mathbb{S}_b}}{N_\mathbb{S}}\right)^2} \notag \\
  =&\sum_{b=0,1}{(\frac{N_{\mathbb{S}_b}}{N_\mathbb{S}}\sum_{i\in \mathbb{S}_b} t^{(i)2}-\frac{1}{N_\mathbb{S}}(\sum_{i\in \mathbb{S}_b} t^{(i)})^2)}. \label{SWMSE}
\end{align}
Through this transformation, the MSE minimization problem becomes a QUBO problem.

\subsection{Formulation as QUBO Problem}
There are two types of QUBO variables to be optimized
$\boldsymbol{\theta}_F$ and $\boldsymbol{\theta}_X$,
and their elements are denoted by $\theta_{F,j}$ and $\theta_{X,i,c}$, respectively.
First, $\theta_{F,j}$ is a binary variable that represent the generated interaction fingerprints,
and finally, based on the calculated result, we use $\theta_{F,j}$ as $f_j$ and generate a new interaction fingerprint.
$\theta_{X,i,c}$ is an auxiliary binary variable for calculating the SWMSE,
where $\theta_{X,i,c}=1$ indicates that
there are $c$ base fingerprints that sample $i$ does not satisfy.
Note that $\theta_{X,i,0}=1$ indicates that sample $i$ satisfies all the producted fingerprints,
i.e., $\theta_{X,i,0}$ indicates the splitting result.
Here, $c$ is an integer from 0 to $M$,
where $M$ is the maximum number of producted fingerprints to be used for generating interaction fingerprints,
and is a learning parameter.
The number of the producted fingerprints that samples $i$ do not satisfy,
denoted as $\text{unsatisfied\_fingerprint}_{i}$,
is expressed by the following equation:
\begin{equation*}
  \text{unsatisfied\_fingerprint}_{i}=\sum_{j}\left(1-\bm{X}_{i,j}\right)\theta_{F,j}.
\end{equation*}
If $\text{unsatisfied\_fingerprint}_{i}=0$, then the sample $i$ satisfies all the producted fingerprints.

\textbf{Loss function}.
The moment of $t^{(i)}$ appearing in Equation \eqref{SWMSE} can be expressed
by using QUBO variables as in the following equations:
\begin{align*}
  \sum_{i\in \mathbb{S}_1} t^{(i)n} &= \sum_{i} \theta_{X,i,0}t^{(i)n} \;\;\;\;\;\;\;\;\;\;\;(n=1,2,..),\\
  \sum_{i\in \mathbb{S}_0} t^{(i)n} &= \sum_{i} (1-\theta_{X,i,0})t^{(i)n} \;\;\;(n=1,2,..).
\end{align*}
Similarly, the numbers of sample groups, $N_{S_1}$ and $N_{S_0}$, that are split by a interaction fingerprint can be written as follows:
\begin{align*}
  N_{\mathbb{S}_1} &= \sum_{i} \theta_{X,i,0}\\
  N_{\mathbb{S}_0} &= \sum_{i} (1-\theta_{X,i,0}).
\end{align*}
As a result, the SWMSE can be expressed as a problem in QUBO form
via the following equation:
\normalsize
\begin{align*}
  \text{SWMSE}&=(\sum_{i}\theta_{X,i,0} t^{(i)2})(\sum_{i}\theta_{X,i,0})\\
  &-(\sum_{i}\theta_{X,i,0} t^{(i)})^2\\
  &+(\sum_{i}(1-\theta_{X,i,0}) t^{(i)2})(\sum_{i}(1-\theta_{X,i,0}))\\
  &-(\sum_{i}(1-\theta_{X,i,0} t^{(i)}))^2.
\end{align*}
\normalsize

\textbf{Constraint function}.
There are three types of constraints on QUBO variables,
as represented by Equation \eqref{cnstr1}-\eqref{cnstr3} below.
Here, Equation \eqref{cnstr1} is a constraint on the relationship between
$\theta_{F,j}$ and $\theta_{X,i,c}$ for each sample.
Satisfaction of this constraint indicates that $\theta_{X,i,c}$ can represent
the number of satisfied base fingerprints used for interaction fingerprints.
Equation \eqref{cnstr2} is a constraint on the validity of $\theta_{X,i,c}$ for each sample.
Equation \eqref{cnstr3} is an optional constraint for narrowing down the search space.
Though $M$ can take values up to $N_B$, we can limit it for practical purposes.
\begin{gather}
  \forall i \sum_{j} (1-\bm{X}_{i,j})\theta_{F,j}-\sum_{c}c\theta_{X,i,c}=0 \label{cnstr1}\\
  \forall i \sum_{c} \theta_{X,i,c}=1 \label{cnstr2}\\
  1\leq\sum_{j} \theta_{F,j} \leq M \label{cnstr3}
\end{gather}
When all these constraints are satisfied,
the actual SWMSE is equal to the SWMSE being calculated in the annealing machine.

\textbf{Hamiltonian}.
The splitting method is explored by minimizing the Hamiltonian given below in Equation \eqref{finalH}.
In this equation, $C_1,C_2$ and $C_3$ are the Hamiltonians for
the constraints expressed in Equations \eqref{cnstr1}-\eqref{cnstr3}.
Note that not all these constraints must be satisfied.
In other words, even if there is a violated constraint,
an estimator can be created using obtained $\theta_{F,j}$.
\begin{align}
  H&=\frac{1}{N_S}\text{SWMSE}(\boldsymbol{X},\boldsymbol{t}|\boldsymbol{\theta}_F,\boldsymbol{\theta}_X)
  +\frac{1}{N_S}C_1(\boldsymbol{\theta}_F,\boldsymbol{\theta}_X)\notag \\
  &+\frac{1}{N_S}C_2(\boldsymbol{\theta}_X)+C_3(\boldsymbol{\theta}_X). \label{finalH}
\end{align}

\begin{table}
  \centering
  \caption{number of the trials that generated effective fingerprints.}\label{tab:tab1}
  \begin{tabular}{|c|cccc|}
    \hline
     &M=2&M=3&M=4&M=5 \\
     \hline
    $N_\mathbb{S}$=50&0&2&2&2\\
    $N_\mathbb{S}$=100 &0&0&0&3\\
    $N_\mathbb{S}$=200 &0&3&0&1\\
    \hline
  \end{tabular}
\end{table}

\begin{figure}
  \centering
  \includegraphics[width=0.8\linewidth,page=3]{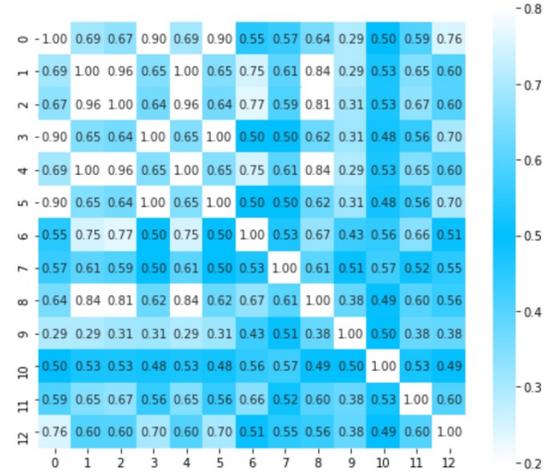}
  \caption{The percentage of matched samples between the generated interaction fingerprints are shown.
  The value of 0 or 1 indicates that two fingerprints substantially the same fingerprints.
  }\label{fig:Bigdata5}
\end{figure}

\begin{table*}
  \centering
  \caption{obtained effective interaction fingerprints.}\label{tab:tab2}
  \begin{tabular}{|c|c|c|c|c|c|}
    \hline
     ID&$N_S$&M&U&I&generated interaction fingerprint \\
     \hline
    1&50&3&3&1756&$\lnot$ N-O$\land$$\lnot$ C\$=C(\$A)\$A$\land$$\lnot$ QCH3\\
    \hline
    2&50&3&3&821&$\lnot$ NC(O)O$\land$$\lnot$ ACH2N$\land$$\lnot$ FRAGMENTS\\
    \hline
    3&50&4&3&1689&$\lnot$ NC(O)O$\land$$\lnot$ QHAQH$\land$$\lnot$ ACH2N\\
    \hline
    4&50&4&2&1027&RING$\land$$\lnot$ QCH3\\
    \hline
    5&50&5&2&151&$\lnot$ NC(O)O$\land$$\lnot$ ACH2N\\
    \hline
    6&50&5&2&116&RING$\land$$\lnot$ QCH3\\
    \hline
    7&100&5&5&2793&$\lnot$ 7M RING$\land$$\lnot$ NAAN$\land$$\lnot$ ACH2N$\land$
      $\lnot$ N=A$\land$$\lnot$ X (HALOGEN)\\
    \hline
    8&100&5&4&2660&$\lnot$ N-O$\land$$\lnot$ QCH2Q$\land$$\lnot$ NC(O)N$\land$$\lnot$ ACH2AACH2A\\
    \hline
    9&100&5&3&1900&$\lnot$ C=CN$\land$$\lnot$ QNQ$\land$$\lnot$ ACH2QH\\
    \hline
    10&200&3&3&1478&NCO$\land$$\lnot$ NC(O)N$\land$$\lnot$ AROMATIC RING\\
    \hline
    11&200&3&3&3399&$\lnot$ NC(O)N$\land$$\lnot$ CH3CH2A$\land$$\lnot$ 6M RING\\
    \hline
    12&200&3&3&2844&$\lnot$ HETEROCYCLIC ATOM$>$1(\&...)$\land$
      $\lnot$ CH2QCH2$\land$$\lnot$ O=A$>$1\\
    \hline
      13&200&5&6&3452&
    $\lnot$ N-O$\land$$\lnot$ NC(O)N$\land$$\lnot$ C=C(C)C$\land$$\lnot$ A\$A!A\$A$\land$
    $\lnot$ A!O!A$\land$$\lnot$ Anot\%A\%Anot\%A\\
    \hline
  \end{tabular}
\end{table*}

\section{Experiment and Discussion}
We evaluated whether QUBO decision trees found effective fingerprints
using multiple fingerprints.
Effective fingerprints mean that they reduce prediction error
comparing minimum one using only one fingerprint.
We used QM9 dataset \cite{Ramakrishnan2014} and predicted value was dipole moment.
Input values were generated based on MACCS keys.
The experiments were performed on a GPU using momentum annealing.
with various $N_\mathbb{S}$ and $M$.
Ten trials were conducted for each parameter.
To see the impact of $M$,
the samples used with the same $N_\mathbb{S}$ were unified.

Table \ref{tab:tab1} lists the number of the trials that generated effective fingerprints.
In case that $M=2$, no effective interaction fingerprints were generated.
Table \ref{tab:tab2} lists the obtained effective interaction fingerprints.
$I$ is the feature importance of each fingerprint when trained using all samples.
Roughly speaking, the larger $N_\mathbb{S}$ and $U$, the higher $I$.
If $U$ is low, $I$ is tends to be low as well
because of its high correlation with the base fingerprints.
If $N_\mathbb{S}$ is low, $I$ tends to be low as well
because of instability associated with small sample size.
Fig. \ref{fig:Bigdata5} shows the percentage of matched samples between the generated interaction fingerprints.
It can be seen that some similar fingerprints are generated,
but overall, different fingerprints are generated for each of them.

We discuss computational cost comparing with a naïve full search method.
The number of combinations to be searched is ${}_{N_\mathbb{F}}C_M$.
When $M=3$ and $N_\mathbb{S}=50$,
calculation time is over three hours a naïve full search model in our environment,
and it shows it is difficult to search fully when $M$ is over 4.
Our experiments using the proposed method were completed within two hours each,
even at $M=5$, suggesting a computational time advantage of our method.

\section{Conclusion}
In this study, we attempted to find effective fingerprints
for determining the chemical property using QUBO decision trees.
As a result, we found interaction fingerprints using multiple fingerprints
reducing prediction error.
Considering the method to utilizing generated interaction fingerprints
and to improve the performance are issues in the future.

\vspace{12pt}

\end{document}